# Benchmarking the Quality of Diffusion-Weighted Images

Jan Klein[1], Sebastiano Barbieri[1], Miriam H.A. Bauer[2], Christopher Nimsky[2], Horst K. Hahn[1]

[1]Fraunhofer MEVIS, Bremen, Germany,
[2]Department of Neurosurgery, University of Marburg, Marburg, Germany

**Abstract**
We present a novel method that allows for measuring the quality of diffusion-weighted MR images dependent on the image resolution and the image noise. For this purpose, we introduce a thresholding technique so that noise and the signal can automatically be estimated from a single data set. Thus, no user interaction as well as no double acquisition technique, which requires a time-consuming proper geometrical registration, is needed.
As a coarser image resolution or slice thickness leads to a higher signal-to-noise ratio (SNR), our benchmark determines a resolution-independent quality measure so that images with different resolutions can be adequately compared.
To evaluate our method, a set of diffusion-weighted images from different vendors is used. It is shown that the quality can efficiently be determined and that the automatically computed SNR is comparable to the SNR which is manually measured in a selected region of interest.

## 1. Introduction

Assuming the real and imaginary data, with mean $A_R$ and $A_I$, corrupted by zero mean Gaussian noise with standard deviation $\sigma$, the probability density function (PDF) of the magnitude data is shown to be a Rician distribution [1, pp. 138-139].
This distribution tends to a Rayleigh distribution for a low signal-to-noise ratio (SNR) and tends to a Gaussian distribution at high SNR. Let $M$ denote the pixel variable of the magnitude image and $A = \sqrt{A_R^2 + A_I^2}$. From the second moment

$$E[M^2] = 2\sigma^2 + A^2 \qquad (1)$$

of the Rician PDF, the noise variance $\sigma^2$ can be determined by estimating $E[M^2]$ from a spatial average of the squared background data points, where $A$ is known to be zero [2, 3, 4, 5, 6, 9]:

$$\hat{\sigma}^2 = \frac{1}{2} <M^2>. \qquad (2)$$

Alternatively, the noise can be measured as the standard deviation of the signal magnitude in air, multiplied by a correction factor $f_e$ of 1.53 [2, 8]. However, both approaches require user interaction which also may introduce errors when regions containing motion artifacts are selected [7].
To overcome these problems, the double acquisition method proposed by Sijbers et al. [6, 9] can be used to estimate the noise if two images are acquired under identical imaging conditions. However, although sometimes two or more MR images are generated for averaging during the acquisition process, these single images are only rarely available to end users. Thus, an automatic detection of noise and signal is shown in the next section.

## 2. A Variance-Based Threshold

Given a single magnitude image $M_j$ of a MR data set consisting of $n$ images, we can use a simple threshold $t$ to cut off all pixel values that are above $t$. That means, if a pixel value $M_j(x,y) > t$, then it is set to zero. The smaller $t$, the more pixels are set to zero. The resulting image which contains only pixel values $0 \leq M_j(x,y) \leq t$ is denoted as $M_j(t)$. A manually fine-tuning of $t$ would allow for separating the object from the background data so that the image noise $\sigma_{M_j^+}(t)$ can be determined from the *positive* pixels (i.e., where pixel values > 0) using the standard deviation of the signal magnitude [2, 8]:

$$\sigma_{M_j^+}(t) = f_e \sqrt{\frac{1}{p} \sum_{j|M_j(x,y)>0} (M_j(x,y) - \mu_{M_j^+})^2} \qquad (3)$$

where $p$ denotes the number of positive pixels and $\mu_{M_j^+}$ their mean value of the image $M_j$.

Averaging the resulting noise values for all $n$ slices of the data set gives a good approximation of the real noise:

$$\sigma_{M^+}(t) = \frac{1}{n} \sum_{j=1}^{n} \sigma_{M_j^+}(t). \qquad (4)$$

However, the manual fine-tuning of the threshold shows to be an error-prone and time-consuming step. If $t$ is chosen too large, then the estimated noise would be too large, otherwise pixel values belonging to the noise would be removed resulting in a holey background and an underestimation of the noise.

Thus, we propose the following variance-based estimation of the threshold. The basic idea is to determine a threshold $t$ so that all images $M_j(t)$ of the data set are as *homogenous* as possible among each other under the condition that no holes appear in the background. Then, the images will contain, as far as possible, those pixels that represent the noise. In the following, we will explain the idea in more detail.

The homogeneity of the images $M_j(t)$ can be measured as the variance of the noise values $\sigma_{M_j}(t)$:

$$\tilde{\sigma}_M^2(t) = \frac{1}{n} \sum_{j=1}^{n} (\sigma_{M_j}(t) - \sigma_M(t))^2 \qquad (5)$$

where $\sigma_{M_j}(t)$ denotes the standard deviation of the image $M_j(t)$ including non-positive pixels and $\sigma_M(t)$ the corresponding average

$$\sigma_M(t) = \frac{1}{n} \sum_{j=1}^{n} \sigma_{M_j}(t). \qquad (6)$$

It is obvious that, for small $t$, $\tilde{\sigma}_M^2(t)$ increases when $t$ increases: pixel values appear more or less randomly in the background when $t$ increases. As a consequence, the variance $\tilde{\sigma}_M^2(t)$ also increases. However, if $t$ becomes larger, the holes in the background are filled more and more so that the variance $\tilde{\sigma}_M^2(t)$ begins to decrease. If the holes are optimally filled, $\tilde{\sigma}_I^2(t)$ assumes its local minimum.

The situation is illustrated in Figure 1 (left) where the variance $\tilde{\sigma}_M^2(t)$ is plotted for a $B_0$ image of a diffusion tensor imaging (DTI) data set that has been acquired on a 1.5T GE scanner. As a consequence, the threshold can be determined as

$$\tilde{t}_{opt} = t \mid \{ t \in [t_1, t_{max}] \wedge t_1 \in \Re \wedge \tilde{\sigma}_M^2(t_1) - \tilde{\sigma}_M^2(t_1 + \varepsilon) > 0 \wedge \varepsilon \in \Re^+ \wedge$$
$$\tilde{\sigma}_M^2(t) \leq \tilde{\sigma}_M^2(\tilde{t}) \ \forall \ \tilde{t} \in [t_1, t_{max}] \} \qquad (7)$$

where $t_{max}$ denotes the maximal pixel value occurring in an image of the data set. The restriction $\tilde{\sigma}_M^2(t_1) - \tilde{\sigma}_M^2(t_1 + \varepsilon) > 0$ ensures that the minimum is not chosen left from the local maximum (in Figure 1 (left), $t_1$ would be at least $\approx 63$).

Note that Equation 7 could be invalid, if the image contains no object but only background pixels. Then, there will be a local maximum of $\sigma_M(t)$ with $\sigma_M(t) > \sigma_M(t_1)$ and $t < t_1$ while at the same time $\tilde{\sigma}_M^2(t)$ could assume a local minimum, see Figure 1 right (for real DTI data

$\sigma_M(t)$ is a continuously increasing function). Fortunately, the problem can easily be solved by an additional condition that the average noise at $t$ must not be larger than at $t_{max}$

$$t_{opt} = t \mid \{t \in [t_1, t_{max}] \land t_1 \in \Re \land \tilde{\sigma}_M^2(t_1) - \tilde{\sigma}_M^2(t_1 + \varepsilon) > 0 \land \varepsilon \in \Re^+ \land$$
$$\tilde{\sigma}_M^2(t) \leq \tilde{\sigma}_M^2(\tilde{t}) \ \forall \ \tilde{t} \in [t_1, t_{max}] \land \sigma_M(t) \leq \sigma_M(t_{max}) \}. \tag{8}$$

Overall, the noise of the data set can be determined from the images $M_j(t_{opt})$ as $\sigma_{M^+}(t_{opt})$. Analogously, the signal can be measured as the mean of the object pixels. In the next section we will describe an efficient algorithm for computing $t_{opt}$.

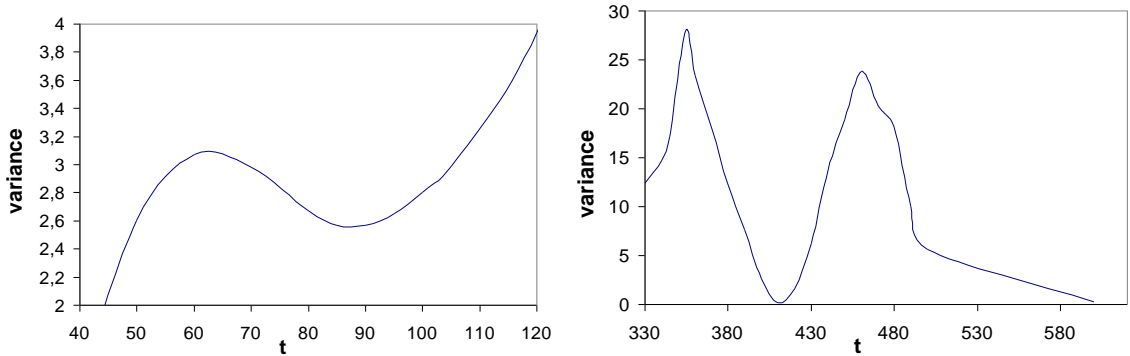

Figure 1: Left: the variance $\sigma_M^2(t)$ assumes its local minimum when holes are filled in the background while simultaneously no pixels from the object are selected. The corresponding threshold $t$ can be used to estimate the noise variance $\sigma_t^2$ of the data set. Right: complex Gaussian noise [10, 11] with sigma=100 added to an image with unique pixel value of 400. As no object is contained in the image, the local minimum at $t \approx 410$ may not be used as $t_{opt}$, instead $t_{opt}=t_{max}$.

## 3. Efficient Threshold Calculation

The naïve computation of $t_{opt}$ would be to test each single $t$. However, this could take a lot of time which would make the overall method less interesting for clinical settings. Let $k$ denote the spectrum of values at which $t$ can assume the minimum of $\tilde{\sigma}_M^2(t)$. Then, the running time for the computation of $t_{opt}$ would be in O($k \cdot n$). Although in our examples $k$ shows to be < 70, we cannot guarantee this value to be an upper bound. As a consequence, it is important that the computation of $t_{opt}$ depends only weakly on $k$ to ensure an efficient computation of the benchmark.

The value of $t_1$ can be determined by a simple heuristic in time O($n$), e.g., starting at $t=40$ and $\varepsilon = 10$, determine $\tilde{\sigma}_M^2(t_1)$ and $\tilde{\sigma}_M^2(t_1 + \varepsilon)$. If its difference is larger than zero, set $t_1=t$, otherwise increase $t$ by 10 and repeat the process. During the computation, $t_{max}$ can be determined at no extra cost.

Afterwards, we propose to use a binary search [12] to find the local minimum in time O(log $k$). Overall, $t_{opt}$ can be determined in time O($n$ log $k$).

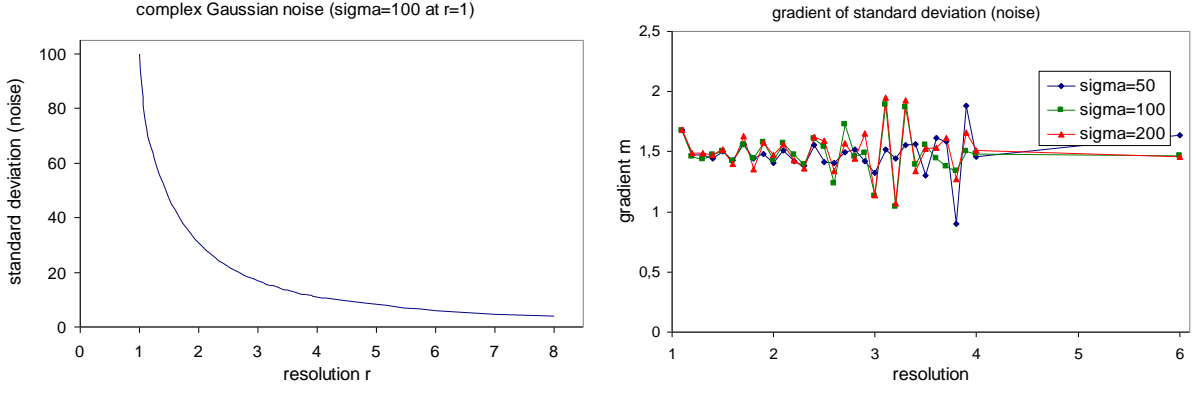

Figure 2: Left: noise, measured as standard deviation [2], depending on image resolution for artificial MR data set. Right: *m*=1.5 shows to be a good approximation for the gradient of the standard deviation.

## 4. Resolution-Independent Quality Measure

It is a well-known fact that the noise of an image decreases when its resolution becomes coarser by averaging its pixels. Let *N* denote the number of voxels that are merged to a single voxel. Then, the noise decreases with $1/\sqrt{N}$ and the SNR increases with $\sqrt{N}$ (law of large numbers):

$$\text{SNR} \propto \sqrt{N} \propto \sqrt{s_v^3}$$

$$\Rightarrow \text{noise} \propto s_v^{-\frac{3}{2}} \qquad (9)$$

where $s_v$ denotes the voxel volume. In MR imaging, equation (9) is not quite correct, because the magnitude data is Rayleigh distributed for high noise. Moreover, if the image contains object pixels and these object pixels are merged by background pixels, the resulting *partial volume effects* affect the correctness of Equation 9 for MR images. Additionally, the used filter for merging (only a nearest neighbor filter would be valid for Equation 9) and non-integer values of *N* also lead to side effects affecting the correctness of Equation 9. Nevertheless, we tested the approximation quality of Equation 9 for MR images. For that purpose, (9) could be written as

$$noise \cong y_0 \cdot s_v^{-\frac{3}{2}}. \qquad (10)$$

Here, $y_0$ denotes the standard deviation (noise) for $s_v$=1.
Logarithmizing Equation 10 shows that log(noise) depends linear on log (resolution):

$$\log(noise) \cong \log(y_0) - \frac{3}{2}\log(s_v). \qquad (11)$$

As a consequence, for testing the approximation quality of equation (9), we have to test whether the gradient m can be approximated by 3/2:

$$m := \frac{\log(noise(r_1)) - \log(noise(r_2))}{\log(r_2) - \log(r_1)} \qquad (12)$$

with $r_1, r_2 \in \Re$, $r_1 < r_2$, and noise($r_1$) denotes the noise at resolution $r_1$. In the next section we will examine *m* for artificial and real diffusion-weighted images.

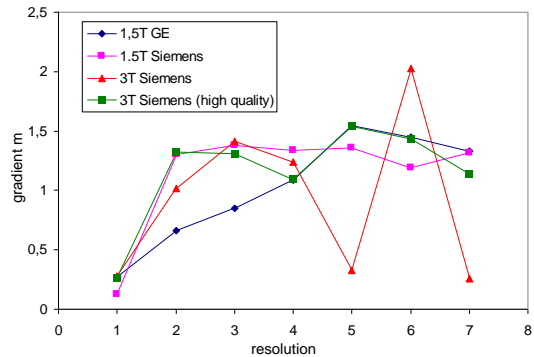

Figure 3: Gradient *m* for real MRI data sets.

## 5. Results

In a first step, we generated artificial MR data sets with high noise ($\sigma$=50, 100, 200) using complex Gaussian noise [10,11] and measured the noise depending on the image resolution. For averaging the voxels, we used a three-lobed Lanczos filter proposed for smoothing DTI images [11]. For an initial noise $\sigma$=100, the image noise depending on the resolution can be found in Figure 2 (left). Figure 2 (right) shows the gradient *m* depending on the resolution for all three artificial data sets ($\sigma$=50, 100, 200). As one can see, *m*=1.5 shows to be a good approximation for the gradient so that Equation 10 shows to be a good approximation for our artificial MR images.

We also measured the gradient *m* for real MR data. For that purpose, five different DTI data sets have been acquired on different scanners (GE 1.5T, Siemens 1.5T, Siemens 3T). For measuring the noise, only $B_0$ images were considered. The results can be found in Figure 3. For most data sets and resolutions larger than 2.0mm, *m*=1.5 is still a good approximation.

## 6. Conclusion and Future Work

We presented a first attempt to allow for measuring the quality of diffusion-weighted images based on a simple threshold technique independent of the resolution. Noise and signal can automatically be computed without any user interaction.

In future work it would be interesting to detect noise which is not based on evenly distributed Gaussian noise. Furthermore, parameters like contrast and sharpness of edges could be utilized for extending the benchmark.